\documentclass[conference]{IEEEtran}
\usepackage{times}

\usepackage[numbers]{natbib}
\usepackage{multicol}
\usepackage[bookmarks=true]{hyperref}
\usepackage{amsmath} 
\usepackage{amsfonts}
\usepackage{bbm}
\usepackage{graphicx}
\usepackage{booktabs}

\makeatletter
\let\MYcaption\@makecaption
\makeatother
\usepackage[font=footnotesize]{subcaption}
\makeatletter
\let\@makecaption\MYcaption
\makeatother\usepackage{graphicx}

\newcommand{\R}{\mathbb{R}}

\newcommand{\ie}{\textit{i}.\textit{e}.}
\newcommand{\eg}{\textit{e}.\textit{g}.}

\DeclareMathOperator*{\argmin}{arg\,min}
\DeclareMathOperator*{\argmax}{arg\,max}
\DeclareMathOperator*{\anash}{a*}

\begin{document}

\title{A Bayesian Framework for Nash Equilibrium Inference in Human-Robot Parallel Play}

\author{Shray Bansal, Jin Xu, Ayanna Howard, Charles Isbell\\
Georgia Institute of Technology\\
\small{Email: \{\texttt{sbansal34, jxu81, ah260}\}\texttt{@gatech.edu}, \texttt{isbell@cc.gatech.edu}}
}

\maketitle

\begin{abstract}
We consider shared workspace scenarios with humans and robots acting to achieve independent goals, termed as parallel play. We model these as general-sum games and construct a framework that utilizes the Nash equilibrium solution concept to consider the interactive effect of both agents while planning. We find multiple Pareto-optimal equilibria in these tasks. We hypothesize that people act by choosing an equilibrium based on social norms and their personalities. To enable coordination, we infer the equilibrium online using a probabilistic model that includes these two factors and use it to select the robot's action. We apply our approach to a close-proximity pick-and-place task involving a robot and a simulated human with three potential behaviors - defensive, selfish, and norm-following. We showed that using a Bayesian approach to infer the equilibrium enables the robot to complete the task with less than half the number of collisions while also reducing the task execution time as compared to the best baseline. We also performed a study with human participants interacting either with other humans or with different robot agents and observed that our proposed approach performs similar to human-human parallel play interactions. 
The code is available at \href{https://github.com/shray/bayes-nash}{https://github.com/shray/bayes-nash}
\end{abstract}

\IEEEpeerreviewmaketitle

\section{Introduction}

People often perform activities in shared spaces with other people achieving their own individual goals. This includes driving to work while sharing the road with other cars, navigating around other shoppers when pushing a cart in a grocery store, and sharing counter-space and utensils in a kitchen. 
Although, these situations are neither purely collaborative nor competitive,
however, the actions of other participants have bearing on each person's own success or failure. We refer to these activities as \textit{parallel play}, related to its psychology namesake that refers to activities in early social development, where children play \textit{besides} instead of \textit{with}, other children~\cite{parten1932social, park2010understanding}. In the Human-Robot Interaction (HRI) context, we define \textit{parallel play} to refer to those activities where people and robots have separate individual goals but interact due to shared space. 
We aim to derive a framework that helps a robot plan effectively for parallel play with human participants, and apply it to a close-proximity pick-and-place scenario between a robot and a human.  

Planning a robot's action in HRI usually involves considering the robot's goal as well as predictions of future human actions~\cite{sadigh2016planning, bansal2018collaborative, koppula2015anticipating}. When working with others, people are often considerate of their intents and beliefs due to Theory-of-Mind~\cite{premack1978does, engel2014reading}, and so, the human's action is influenced by their predicted plans of the other participant's, including the robot. Modeling this cyclical-dependence, of the human's predicted plan on the robot's and vice-versa, is important for accurately representing the interaction dynamics in HRI. 

Game Theory provides us tools to model this inter-dependence of rational interacting agents. The Nash equilibrium (NE) is a set of actions, one for each agent in the game, which is optimal,  assuming the actions of others remain fixed~\cite{leyton2008essentials}. A Nash equilibrium implicitly captures the inter-dependence between agents, and our approach plans by finding equilibria to enable better coordination. 

Although humans have been shown to play to Nash equilibrium~\cite{mailath1998people},
we find that multiple equilibria can exist in a game and it is not clear how the robot should choose between them. Figure \ref{fig:picking_together} shows an example of two equilibria in a pick-and-place scenario, where each favors a different agent by allowing them to reach their goal first. 
A collision is likely if both agents choose an equilibrium that favors them, highlighting the importance of coordination.
Humans can coordinate social behavior in non-competitive games by learning and following social norms~\cite{ho2016feature}. Here, a norm refers to a set of abstract instructions that agents follow, and expect others to follow. 
In Figure \ref{fig:picking_together}, a norm might favor solutions that allow the agent closer to their goal to reach for it first and, if followed by both agents, would lead to coordination. However, sometimes agents can ignore the norm in favor of their personal preferences. For example, a selfish agent might ignore the norm and expect the other to always yield to them when reaching for an object. Coordinating with such agents requires the ability to infer this preference from experience. 

\begin{figure*}[]
  \centering
  \begin{subfigure}{.72\textwidth}
    \includegraphics[width=.95\linewidth]{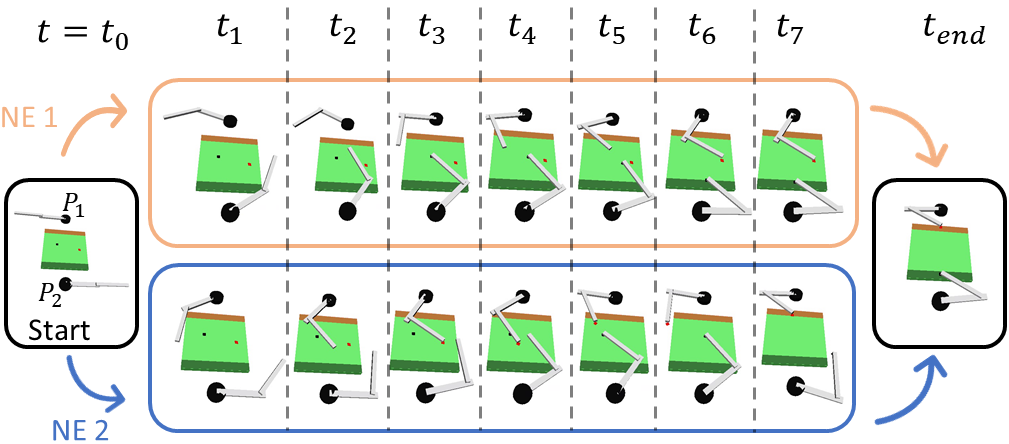}
    \caption{Nash Equilibirum (NE) Action Sequences}
    \label{fig:picking_together}
  \end{subfigure} 
  \begin{subfigure}{.26\textwidth}
    \includegraphics[width=.99\linewidth]{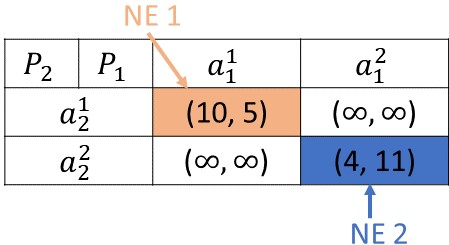}
    \caption{Cost Matrix}
    \label{fig:individual_sub}
  \end{subfigure}
  \caption{A scenario where two agents, $P_1$ and $P_2$, move the red and black blocks to their respective goal locations. In (a), we show two sequences of actions starting in the same state at $t_0$, and ending in the same goal state at $t_{end}$. While each action sequence (trajectory) is a Nash Equilibrium (NE), they both favor different agents. NE1 (top) favors $P_2$ by allowing it to reach its goal first, NE2 (bottom) favors $P_1$. A cost function is illustrated in (b), where each cell is a tuple containing the cost of the actions for $P_1$ and $P_2$ respectively. Here, our action space has only $2$ actions per agent, $A_1= \{a^1_1, a^2_1\}$ and $A_2= \{a^1_2, a^2_2\}$. 
  }
    \label{fig:picking_together}
\end{figure*}

We design a framework that finds Nash equilibria for \textit{parallel play} tasks; it models the strategy for choosing equilibria as a distribution composed of two aspects - (1) a domain-specific social norm, designed by an expert apriori and (2) an agent-specific individual preference, inferred online during the interaction. We hypothesize that this framework 
would lead to better coordination with humans in performing in such tasks, due to its modeling of the decision-making coupling between agents, as well as, its combination of expert knowledge with online adaptation. 
To validate, we apply this to a close-proximity pick-and-place task, designed to be similar to HRI tasks used to study team coordination and fluency~\cite{gabler2017game, mainprice2016goal} with a simulated human. Our results show that this framework is able to shorten task execution time while also reducing the number of human collisions by half as compared to the best baseline when interacting with a simulated human with $3$ potential personalities. 

We make the following contributions:
\begin{enumerate}
    \item Introduce a novel framework that models norm-following social behavior and personality-based likelihood inference to interactive-planning with humans in parallel play activities. We do this by first computing the Nash equilibria and then using this framework to find a distribution over them.
    \item Design task and metrics to benchmark the performance of interactive-planning algorithms for \textit{parallel play}. This includes $3$ baselines for simulating distinct human personalities that have similarity to human performance of the task, which enables testing adaptability of the algorithms to different human behavior.
\end{enumerate}

\section{Related Work}
There has been extensive work in HRI for planning a robot to work on tasks around humans in domains like parts assembly~\cite{hawkins2014anticipating, gabler2017game}, motion planning, and autonomous driving~\cite{sadigh2016planning, bansal2018collaborative}. Planning around people generally involves two aspects, predicting the human's behavior and finding robot actions that achieve its goal in the presence of the human. 

\textbf{Human modeling.} Prior work has placed emphasis on accurately modeling the human's rational goal-driven behavior. This includes learning the human's preferences to predict low-level trajectories through a reward function obtained by inverse reinforcement learning~\cite{ziebart2009planning, sadigh2016planning} or high-level decision-making and choice of goal or the timing of actions for part delivery using a probabilistic task model~\cite{hawkins2014anticipating, gombolay2015decision}. Some models were also adaptable in modeling the preferences of the particular human with whom the robot was interacting~\cite{nikolaidis2015efficient}. 

\textbf{Human adaptive planning.} A common approach to planning for this interactive setting is by predicting the human's behavior and then finding a best-response to this behavior. This approach works very well in scenarios where the robot assumes an assistive role like parts delivery~\cite{hawkins2014anticipating, unhelkar2014comparative} where, apart from ensuring that the human doesn't wait, the robot intends to avoid interactions by keeping out-of-the-way. It has also been used to plan a robot to pick-up objects in a close-proximity scenario where the robot planned trajectories that did not intersect with predicted human plans~\cite{mainprice2016goal, li2019safe}. An inherent assumption, here, is that, although the human's plan depends on the situation, the prediction is independent of the robot's plan. So, in situations where the agents have equal roles, like driving or navigation, the robot will choose overly conservative behaviors which can lead it to freeze when trying to navigate crowds~\cite{trautman2010unfreezing} or fail to merge in traffic~\cite{sadigh2016planning}.

\textbf{Mutually adaptive planning.} Recent work has addressed this by considering the human's influence-ability as well. Their model includes both the influence of the human and their goals on the robot and the influence of the robot on the human. Similar to us, they also utilize game-theoretic tools to model this cyclical influence. \citet{turnwald2019human} modeled robot navigation as a dynamic general-sum game and computed a Nash equilibrium to effectively plan the robot's trajectory among a crowd of pedestrians. \citet{sadigh2016planning} modeled driving as a Stackelberg game where the robot planned first and the human planned in response; they showed that this model can successfully influence human behavior in simulated driving tasks. \citet{fisac2019hierarchical} extended this to longer time horizons by computing a Nash equilibrium for high-level actions and optimizing low-level trajectories for executing them. \citet{gabler2017game} utilized the Nash equilibrium to find an order for object pick-up in a close-proximity pick-and-place task similar to ours; they found that considering the mutual adaptation allowed their framework to improve safety as well as human subjective preference. Our approach also uses the Nash equilibrium to plan goal-driven actions for the robot that consider the mutual adaptability between the two agents. However, our approach includes a strategy for selecting an equilibrium in case multiple are present, while others either have not mentioned this strategy~\cite{gabler2017game} or only find one equilibrium due to their problem structure~\cite{sadigh2016planning, fisac2019hierarchical}. 

\textbf{Online model inference.} Although different people perform the same task in different ways, prior work uses a single model to describe all human users, with some exceptions. While in~\cite{nikolaidis2015efficient}, Nikolaidis et. al. cluster human behavior into multiple \textit{types} and predict actions based on the inferred \textit{type}, in \cite{nikolaidis2016formalizing}, they group people by their adaptability to the robot's actions. \citet{chen2018planning} explicitly model human \textit{trust} on the robot's ability during decision-making and infer this parameter during the interaction. \citet{sadigh2016information} use information-gathering actions to infer whether a driver is attentive or not and use this to coordinate better. We define a latent variable that represents the human personality and infers this online, for each participant, in order to find plans that coordinate well with the human. Recently, \citet{schwarting2019social} proposed a method for simulated autonomous driving around human drivers using Nash equilibrium. Similar to us, they have a parameter that represents human-personality and infers it online. However, in their model, this parameter is part of the human cost function while we use it to choose between Nash equilibria. Also,  while they define a continuous action-space and use a local approximation for computing Nash equilibrium, which finds a single equilibrium, we find multiple equilibria since we plan discrete high-level actions. 
Their results indicate that inferring human preference in combination with finding Nash equilibrium improves prediction and helps achieve coordination with humans.

\textbf{Coordination. } The importance of coordinating with a human in non-competitive games was highlighted by ~\citet{carroll2019utility} where they learn human models and used them to train reinforcement learning agents that achieve performance superior to self-play. Similar to us, \citet{ho2016feature} proposed using social norms for improving coordination in multi-agent environments. While our approach does not use learning, it can be incorporated into our framework to replace the user-defined costs or norms.

\section{Interactive Planning by Nash Equilibrium}
We model the multi-agent interactive planning task as a non-cooperative game represented as tuple $G$, $G = (P, A, c)$~\cite{leyton2008essentials}. Here, $P = \{P_1, ..., P_N\}$ is a finite set of $N$ players, $A = A_1\times ...\times A_N$ where $A_i$ is the set of actions available to player $i$. We refer to the set of concurrent actions, one for each agent, as an action profile, $a$, $a=(a_1,..., a_N)$. We define a cost representing the unfavorability of an action profile for agent $i$ as $c_i: A \mapsto \R$ and $c = (c_1, ..., c_N)$ includes the mapping for all agents.

In our scenario, each agent $p$ is a robot arm, each action set $A_p$ is a set of goal-driven trajectories, each trajectory is a sequence of joint-space positions and velocities sampled using a planner, and the cost $c_i(a)$ encourages each robot to minimize task completion time while avoiding collisions with other agents. The goal for an agent $i$ is to take an action $a_i \in A_i$ in profile $a$, which minimizes its cost. However, its cost depends upon the actions chosen by the other agents in the profile $a$. We assume that all agents are rational and have Theory-of-Mind, \ie, they choose actions to minimize their own cost and are aware of the states and goals of the other agents. These assumptions allow us to utilize the Nash Equilibrium (NE) solution concept for this game. An action profile is a NE, for a single-stage game, if no agent has an incentive to choose a different action for themselves given that all the other actions are fixed.

\begin{equation}
    \label{eq:nash}
    \anash_i \in \argmin_{a_i} c_i(\anash_1, ., a_i, ., \anash_N ) \quad \forall i \in N.
\end{equation}

Although generally, only one (mixed) equilibrium is guaranteed to exist for a game~\cite{leyton2008essentials}, in our problem, one pure equilibrium is always present and we find that multiple equilibria are frequently present. For the planning agent, some of these equilibria can be eliminated for being Pareto-sub-optimal, \ie, worse for all agents. For example, Nash profile, $\anash^1$ Pareto-dominates a profile $\anash^2$, if $c_i(\anash^1) < c_i(\anash^{2}) \forall i \in N$. Next, we present an approach that can help the agent select an action by choosing between Pareto-optimal equilibria.
 
\section{Optimal Equilibrium Selection}
Our strategy chooses between equilibria using two aspects of human social behavior, norm-following and personality-adaptation. We model the distribution over equilibria as a product of its probability under the norm, $p_{n}$ and its probability given the predisposed personality, $p_{\alpha}$,

\begin{equation}
    \label{eq:distribution_overall}
    p(a) = p_{n}(a) p_{\alpha}(a).
\end{equation}
Here, and in the rest of this section, $a$ refers to a NE action profile. Next, we explain the norm for this problem and how we use observations to update the personality distribution. 

\subsection{Norm}
Similar to~\cite{ho2016feature}, we define a norm to be a set of, situation-dependent, abstract, instructions that agents follow with the expectation that others will follow them as well. They help agents coordinate in the absence of prior knowledge of the agents they are interacting with.
For example, a first-come-first-leave norm can help decide how cars navigate a four-way stop. Here, we model it as a probability distribution over NE. Different games will have different norms and the choice of a norm should be based on expert knowledge or learned from data. For our problem, we use a simple min-norm, that prioritizes the equilibrium which achieves minimum cost for any of the agents, 
\begin{equation}
    \label{eq:norm}
    p_{n}(a) \propto e^{-\lambda_n min_i(c_i(a))},
\end{equation}
where $\lambda_n$ is a parameter of the exponential distribution that we set.
This norm it encourages the agent with the shortest unobstructed path to its goal to act first. 

\subsection{Online preference estimation}

Although norms can help in coordination, people sometimes have strong preferences that guide them towards certain equilibria regardless of the norms. For example, an aggressive driver may decide to cross an intersection first, despite the norm, expecting the other drivers to adapt their strategy. 
We model this as a distribution over equilibria, inferred at time $t$ using the history $H_t$ of the past interaction,
\begin{equation}
    \label{eq:p_personality}
    p_{\alpha}^{t}(a) = p(a | H_t).
\end{equation}

Here, $H_t$ refers to the history of the interaction, \ie, $H_t = \{(\{s^0_{i \in N}\}), ..., (\{s^{t-1}_{i \in N}\})\}$, and $s_i^t$ is the state of agent $i$ at time $t$.
We set it to the uniform distribution at the start,
\begin{equation}
    p_{\alpha}^{t=0}(a) = \text{uniform}(a) \quad \forall a.
\end{equation}

We define an exponential distribution on the distance between a past trajectory, $H$, to an action profile, $a$,

\begin{equation}
    \label{eq:infer1}
    p(a | H) \propto e^{-\lambda_\alpha f_{dist}(a, H)},
\end{equation}
 
where $f_{dist}(a, H)$ is defined as the euclidean distance between the sequence of states in $H$ to those sampled at the same time increments from $a$ and $\lambda_\alpha$ is a parameter of the exponential distribution. Although, we would like to use Eq. \ref{eq:infer1} to compute $p(a^t| H^t)$, however, the distance of past trajectories to current action profiles is not meaningful because the last state, $s_t$, in the history is the first state of all action profiles at time $t$. 
So, we only use Eq. \ref{eq:infer1} to find $p(a^0|H^t)$ and define a latent variable $\theta$ to help infer $p(a^t|H^t)$. We use $\theta$ to represent a personality-based distribution over equilibria, $p(a^t|\theta)$, and assume that it remains constant for every agent during an interaction. 
We derive $p(a^t|H^t)$ by using the personality, $\theta$, 
\begin{gather*}
    p(a^t|H^t) = \sum_{\theta}{p(a^t, \theta| H^t)}, \\
    p(a^t|H^t) = \sum_{\theta}{p(\theta| H^t)} p(a^t| \theta, H^t). 
\end{gather*}

We assume that the personality, $\theta$, encodes the information required for predicting the agent's next action, which makes $a^t$ conditionally independent of $H^t$ given $\theta$. So,
\begin{equation}
    p(a^t|H^t) = \sum_{\theta}p(\theta| H^t) p(a^t| \theta).
\end{equation}

We define $\theta$ such that each action profile $a$ that belongs to a personality is equally likely to be chosen, 
\begin{equation}
    \label{eq:action_theta}
    p(a| \theta) = \frac{\mathbbm{1}_\theta(a)}{\sum_{a'}\mathbbm{1}_\theta(a')}. \\
\end{equation}

Next, we find $p(\theta| H^t)$ by taking its joint distribution with $a^0$ and marginalizing it out,
\begin{gather*}
    p(\theta| H^t) = \sum_{a^0} p(\theta, a^0| H^t), \\
    p(\theta| H^t) = \sum_{a^0} p(a^0| H^t) p(\theta|a^0,H^t).
\end{gather*}

From the conditional independence between $\theta$ and $H^t$ given $a^0$, we get,
\begin{equation}
\label{eq:infer_hist}
p(\theta| H^t) = \sum_{a^0} p(a^0| H^t) p(\theta|a^0).    
\end{equation}

Since we assume a uniform prior on $\theta$, $p(\theta)$ is a constant. Combining with Eq. \ref{eq:action_theta}, we get,
\begin{equation}
    \label{eq:infer_personality}
    p(\theta|a^0) =  \frac{p(\theta) p(a^0|\theta)}{p(\theta)\sum_{\theta'}  p(a^0|{\theta'})} =
    \frac{p(a^0|\theta)}{\sum_{\theta'}  p(a^0|{\theta'})}
\end{equation}

We use $p(\theta|a^0)$ and $p(a^0|H^t)$ (Eq. \ref{eq:infer1}) to get $p(\theta| H^t)$ in Eq. \ref{eq:infer_hist}. This allows us to find $p(a^t|H^t)$ from Eq. \ref{eq:infer_hist} by using $p(a^t|\theta)$ from Eq. \ref{eq:action_theta}, which gives us $p_{\alpha}(a^t)$ in Eq. \ref{eq:p_personality}

\section{Pick-Place Task}
\begin{figure}[h]
  \centering
   \includegraphics[width=0.9\linewidth]{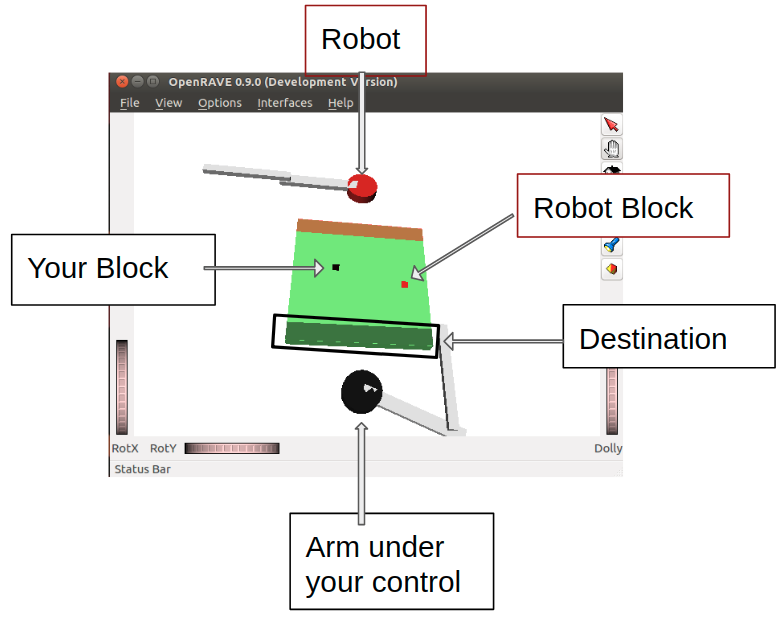}
  \caption{Pick-and-place task scenario.}
    \label{fig:scenario}
\end{figure}

The pick-and-place task involves two $2$-dof articulated arms moving on a 2D surface with the goal to pick up their designated object, by moving their end-effector close to it for grasping, and placing it, by bringing the grasped object to the destination area. The scenario is depicted in Figure \ref{fig:scenario}, where the arm with a red base was controlled by our approach, and the other one was either simulated as a human or controlled by a human. Henceforth, the former will be referred to as the robot and the latter as the human. 

\subsection{Action Planning}
\begin{figure}[h]
  \centering
   \includegraphics[width=0.95\linewidth]{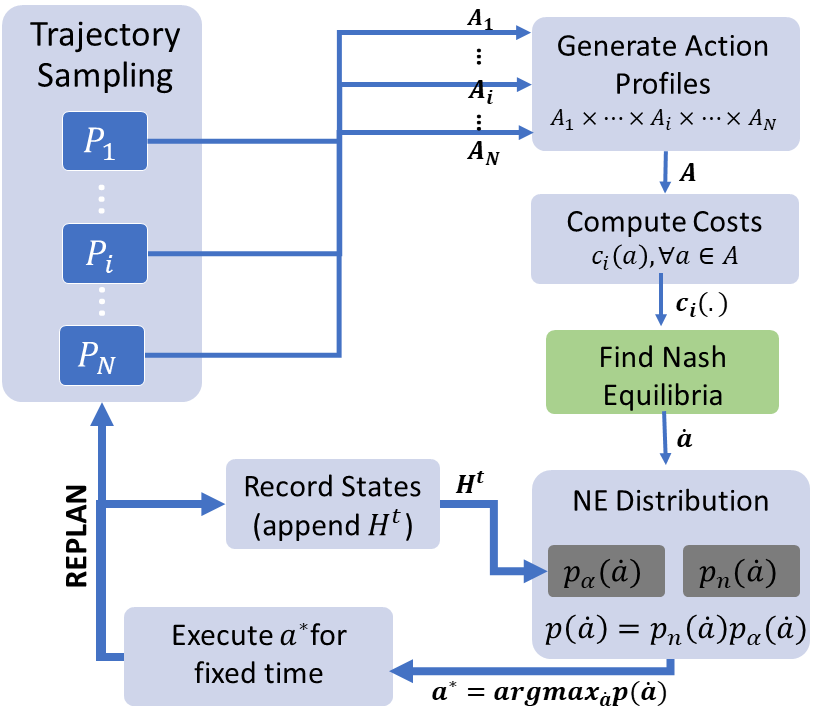}
  \caption{Framework. We first plan trajectories, then use the cost to compute all Pareto-optimal Nash equilibria, then combine the norm and inferred-personality distributions to select the most likely equilibrium. This action is partially executed before repeating the whole process until the task completes.}
    \label{fig:framework}
\end{figure}

To plan for this task, we first sample $k-1$ plans for each agent in configuration space using a Rapidly-exploring Random Tree (RRT)~\cite{lavalle1998rapidly} and add a static plan where the agent does not move, $A_i = \{\tau_{j \in k}\}$. We use them to generate an action set by taking the outer product of the trajectories for each agent, $A = A_1 \times ... \times A_N$. We compute a cost for each action profile by simulating it and use Eq. \ref{eq:nash} to find the Nash equilibria. We choose the NE profile $a$ that maximizes the distribution $p(a)$ from Eq. \ref{eq:distribution_overall}, and select the agent's action from $a$. This action, along with the action taken by the human, is executed until a collision is detected or if the time before replanning is reached. After this, we update the history $H^t$ and replan. This process continues until the robot completes the task and is depicted in Figure \ref{fig:framework}.

\subsection{Task Costs}
We define a simple cost function that encourages the robot to complete the task quickly and avoid collisions. The cost of an action profile, $a = (a_R, a_H)$, where $a_R$, $a_H$, are the robot and human actions respectively, is the trajectory duration if it is successful in reaching the goal and infinity if it leads to a collision. We sample goal-reaching trajectories for each agent that are independent of the goal and state of other agents. We assume that the human is also goal-driven and collision-avoidant and so assume an analogous cost function where their cost depends on the human task completion time. Under these conditions, one pure Nash equilibrium is guaranteed as long as there exists a trajectory for the robot to reach the goal.
For instance, say that we select an action profile with the robot's action being the shortest trajectory to its goal and the human's action as the shortest trajectory that does not collide with the robot's plan (including the static action). This profile will be a Nash equilibrium since neither agent has an incentive to modify their actions. For the robot, the action is optimal, and, for the human, this action is optimal assuming the robot's action as fixed due to the infinite cost of a collision.

\subsection{Baselines}
We define three baselines to compare with our approach.
\begin{enumerate}
    \item Defensive. The robot chooses an action assuming that the human wants to maximize the robot's cost while still achieving its goal leading to a maximin formulation. Thus, the agent will act defensively by preferring actions that do not lead to collision with the sampled human trajectories and will often lead it to wait for the human to complete their task. 
    \begin{equation}
        a_R = \argmin_{a_R \in A_R} \quad max_{a_H \in A_H}c_R(a_R, a_H).
    \end{equation}
    \item Selfish. Chooses an equilibrium profile that minimizes the robot's cost. This strategy selects a trajectory that reaches the goal as quickly as possible assuming the other agent avoids collision.
    \begin{equation}
        a_R \in \anash, \anash = \argmin_{\anash} c_R(\anash).
    \end{equation}
    \item Norm-Nash. Chooses an equilibrium profile that maximizes the norm distribution $p_n$ from Eq. \ref{eq:norm}. This leads to behavior that encourages the agent closer to their goal to reach them first. While the first two will lead to somewhat fixed behaviors, this strategy adapts to goal-achievability, which varies across tasks and also in state evolution within the same interaction. 
    \begin{equation}
        a_R \in \anash, \anash = \argmax_{\anash} p_n(\anash).
    \end{equation}
    
\end{enumerate}

\subsection{Implementation Details}
A 3D simulation environment was created using the open-source Open Robotics Automation Virtual Environment (OpenRAVE)~\cite{diankov_thesis} with a time step of $0.1$ seconds. The action-set, $A_i$ was sampled using an RRT planner from the open-source Open Motion Planning Library (OMPL)~\cite{sucan2012the-open-motion-planning-library}. We sampled $k=8$ plans for each agent when planning and compute the cost as an $k\times k$ table by simulating the actions using OpenRAVE with a time-step of $0.8$; we increased the time-step here to allow for fast computation of the nash solutions. Parameter $\lambda_n$ of the norm distribution (Eq. \ref{eq:norm}) was set to $50$. We set two personality types and use a binary latent variable $\theta=\{0,1\}$. $\theta=0$ selects equilibrium profiles $a$ that favor agent $1$, $c_1(a)< c_2(a)$, and $\theta=1$ selects equilibrium profiles $a$ that favor agent $2$, $c_2(a)< c_1(a)$. We set $\lambda_\alpha =10$ in the personality distribution (Eq. \ref{eq:infer1}).

\section{Simulated Human Study}

We simulate human behavior to create a controlled setting for our first experiment. 

\subsection{Simulated Human} We defined three human behaviors using the baselines: (1) \textbf{Defensive}, (2) \textbf{Selfish-Nash}, and (3) \textbf{Norm-Nash}. We chose the first two behaviors because of their clear intuitive distinctness and combine it with the third in accordance with our expectation that people also follow social norms. 

\subsection{Metrics} We measured the following task performance metrics: total task completion time and task time for each agent; we also counted safety stops, which are the number of times the simulation stopped the agents to avoid an impending collision. To keep these measures independent, we did not have any time penalty for a safety stop. 

\begin{figure}[]
  \centering
   \includegraphics[trim=0 0 0 0, clip, width=1\linewidth]{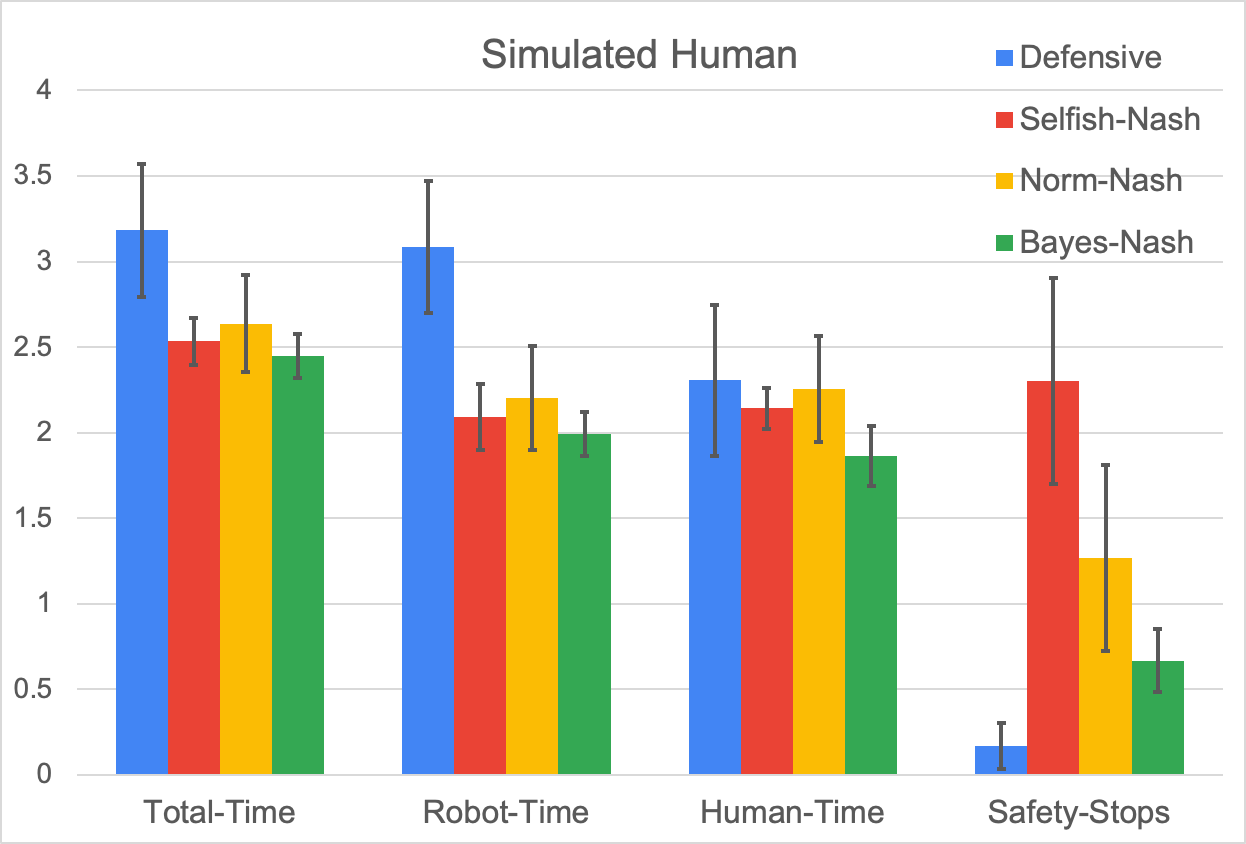}
  \caption{Results from the experiment involving a simulated human. Note that our proposed approach, Bayes-Nash, is safer than all the non-defensive baselines and similar in task completion time to the Selfish-Nash baseline. Error bars represent standard error of the mean (SEM).}
    \label{fig:result_rr}
\end{figure}

\subsection{Results} To test how each algorithm fares with the different behaviors, we randomly pair the robot with one of these simulated human behaviors with random object locations for $30$ trials. The averaged metrics for the three baselines and our proposed approach, Bayes-Nash, are presented in Figure \ref{fig:result_rr}. As expected, the Defensive robot was the safest, but its safe behavior also caused the highest robot and total task completion times. The Selfish-Nash was significantly faster than the Defensive robot but also led to the highest safety stops. Both Norm-Nash and Bayes-Nash performed comparably in time to Selfish-Nash but the Bayes-Nash was marginally faster. They were both significantly safer than Selfish-Nash and Bayes-Nash also had the fewest safety stops of the two.

\subsection{Analysis} These results illustrate the trade-off between safety and efficiency present in the task, where the Defensive and Selfish-Nash agents sit at opposite extremes. Norm-Nash and Bayes-Nash are able to better trade-off these metrics due to their capability to adapt to the situation and the (simulated) human, respectively. Next, we perform an experiment to validate this trade-off in human interaction.

\section{Human-Human Study}

\begin{figure}
  \centering
  \begin{tabular}[b]{c}
    \includegraphics[trim=5 7 10 30, clip, width=.85\linewidth]{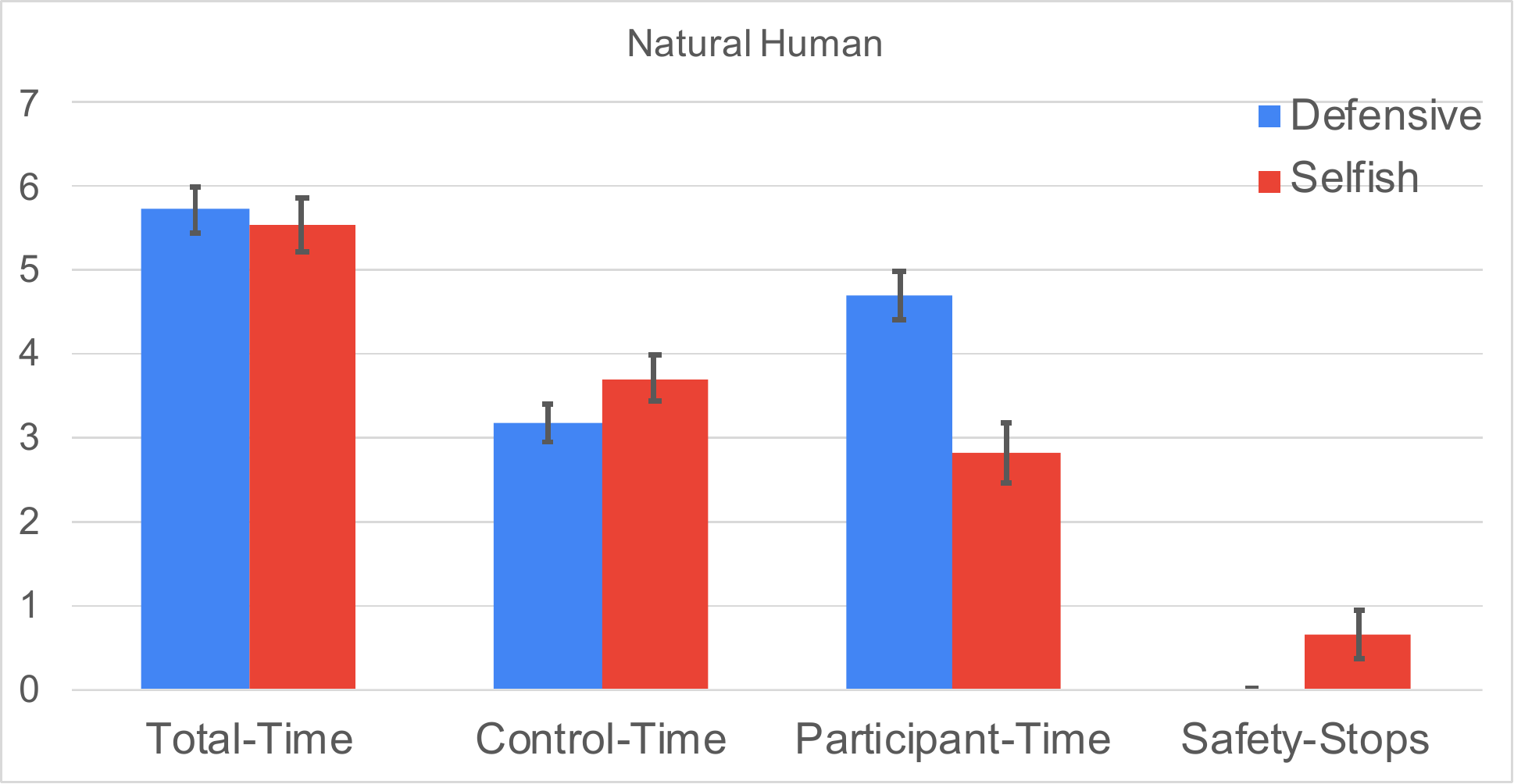} \\
    \small (a) Natural (\textit{Control}) Human
  \end{tabular} \qquad
  \begin{tabular}[b]{c}
    \includegraphics[trim=5 7 10 30,clip,width=.85\linewidth]{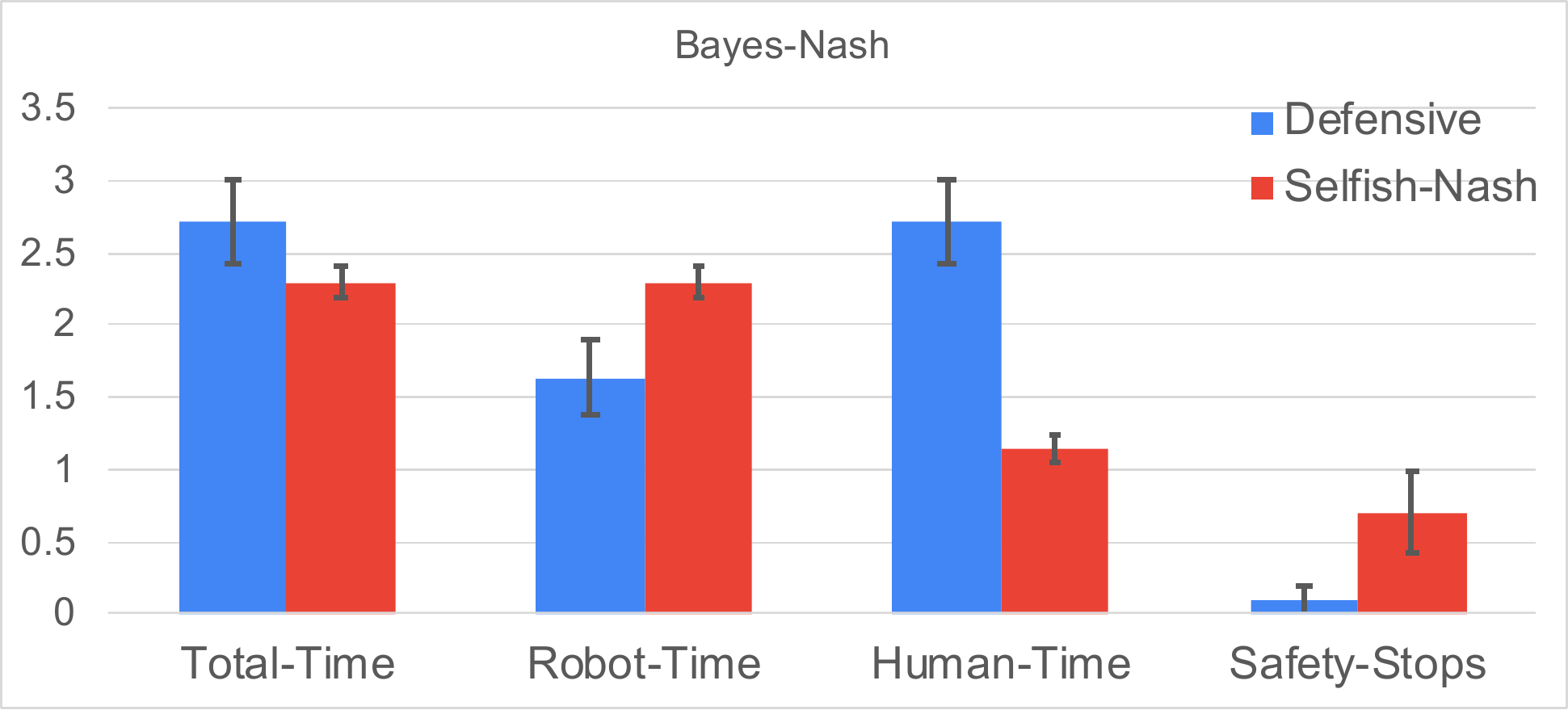} \\
    \small (b) Bayes-Nash
  \end{tabular}
  \caption{ (a) Plots the interaction task metrics for the naturally acting human in the presence of either a selfish or defensive participant in the human-human study; (b) The same metrics but for the interaction of Bayes-Nash with the Selfish and Defensive baselines. The similarity in the relative trends across (a) and (b) highlight the similarity of Bayes-Nash to a real human agent. Error bars represent SEM.}
  \label{fig:results_hh}
\end{figure}

To investigate the natural interaction between two people, we recruited $4$ participants to perform the same task, in a pilot experiment, where both interacting agents were human and controlled the simulated robot arm using a gamepad controller. 

\subsection{Experiment Design} We kept one of the human agents fixed throughout the experiment and will refer to them as \textit{control}. The other agent (participant) evoked different behaviors in each experiment and performed $3$ trials with the control. In the first trial, the participant was asked to behave \textbf{naturally}, by trying to increase efficiency while reducing task time. For the other two trials, the participant chose either (1) \textbf{Selfish} - completing this task efficiently or (2) \textbf{Defensive} - avoiding collisions with the other arm strategy. The \textit{control} kept the same natural strategy throughout the interactions and was not made aware of the strategy that the participant was employing. Also, no verbal communication was allowed during the experiment. We measured the same metrics as in the simulated human experiment.

\subsection{Results} Figure \ref{fig:results_hh} (a) shows that the total task completion times for the Selfish and Defensive humans are similar when interacting with a naturally-acting human. However, in terms of their individual task completion times, the Defensive agent takes significantly longer as compared to the Selfish agent. The Selfish human also triggers more safety stops but the safety stops in this study were much less than in simulation. In Figure \ref{fig:results_hh} (b) we show the simulated Defensive and Selfish-Nash behaviors when interacting with Bayes-Nash. We find similar comparative trends between behavior types for both task completion times and safety stops. However, the robot in (b) completes the task more quickly since the arm is allowed higher velocities in simulation.

\subsection{Analysis} 
These results indicate that a naturally-acting human adapts well to both Selfish and Defensive behavior due to similar task metrics for both conditions, as shown in Figure \ref{fig:results_hh}(a). Similar trends for Bayes-Nash, Figure \ref{fig:results_hh}(b), indicate that it also adapts well to different strategies. 
The similarity between trends among personalities across experiments validate our interactive task design and metrics to benchmark performance for \textit{parallel play}. Also, since the latent variable used to parameterize the equilibrium was designed to capture the agent's favorability in equilibria and not the specific behaviors of the baselines, we expect that Bayes-Nash to be able to adapt well to real human participants. This leads to the following two hypotheses for a human-robot interaction study:

\textbf{H1: } A robot using Bayes-Nash will have significantly fewer collisions than a robot using a Selfish-Nash planner.

\textbf{H2: } A robot using Bayes-Nash will have faster task completion time than a robot using a Defensive planner.

\section{Human-Robot Study}
We test these hypotheses in a pilot experiment by pairing human participants with a robot controlled by our algorithm and the baselines.   
\subsection{Experimental Design}
\begin{figure}[]
  \centering
  \includegraphics[width=0.9\linewidth]{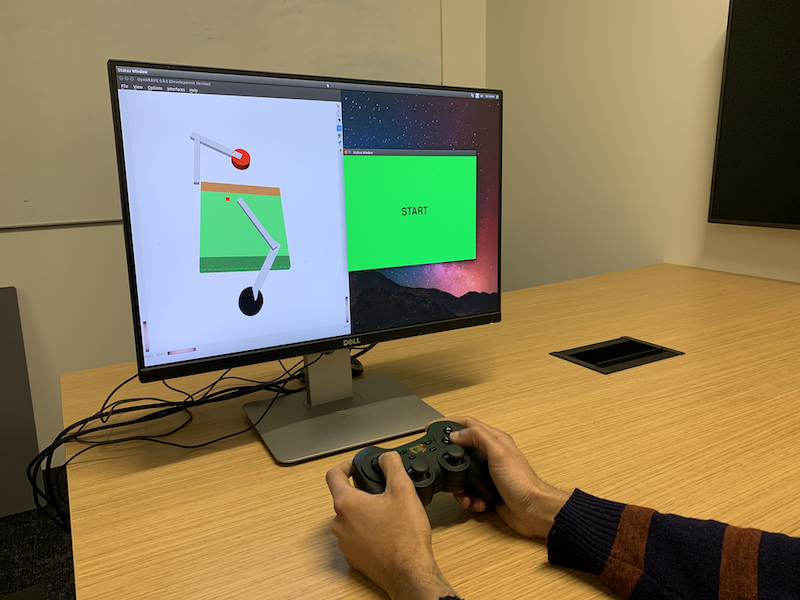}
  \caption{A user controlling the robot arm during the Human-Robot study.}
    \label{fig:exp_robot}
\end{figure}

In order to validate the proposed approach, we design a within-subject human study. We examined the effects of three planning algorithms on their interaction with a human user and counter-balanced their order. Participants were asked to control a robot arm in simulation using a gamepad controller, as shown in Figure \ref{fig:exp_robot}. The gamepad controller allows users to move the robot arm in eight different directions at  $10$ Hz. We used the same manipulation task as the previous two experiments, including keeping the object locations the same. Participants were informed that they might interact with different robots but not what these types were. There were three rounds of the task and each round involved six trials. In each round, the robot used one of the following conditions: (1) \textbf{Defensive}, (2) \textbf{Selfish-Nash}, and (3) \textbf{Bayes-Nash}.
We used the same metrics as before.

\subsection{Procedure}
After giving informed consent, participants went through an overview of the experimental procedures. The study started with a pre-survey to collect demographic information. Then participants entered a practice session to familiarize themselves with the user interface and the gamepad controller. The purpose of this session was to erase potential novelty effects caused by the robot and the user interface. The practice round ended when the participants indicated they felt comfortable with the control and overall task. Then the participants went through three rounds of ‘pick-and-place tasks’ with the robot. Each round consisted of 6 trials and participants were asked to fill out a short survey after the last trial. After these three rounds, the participants were given a post-survey with open-ended questions about their experience.

\begin{figure}[]
  \centering
   \includegraphics[width=1\linewidth]{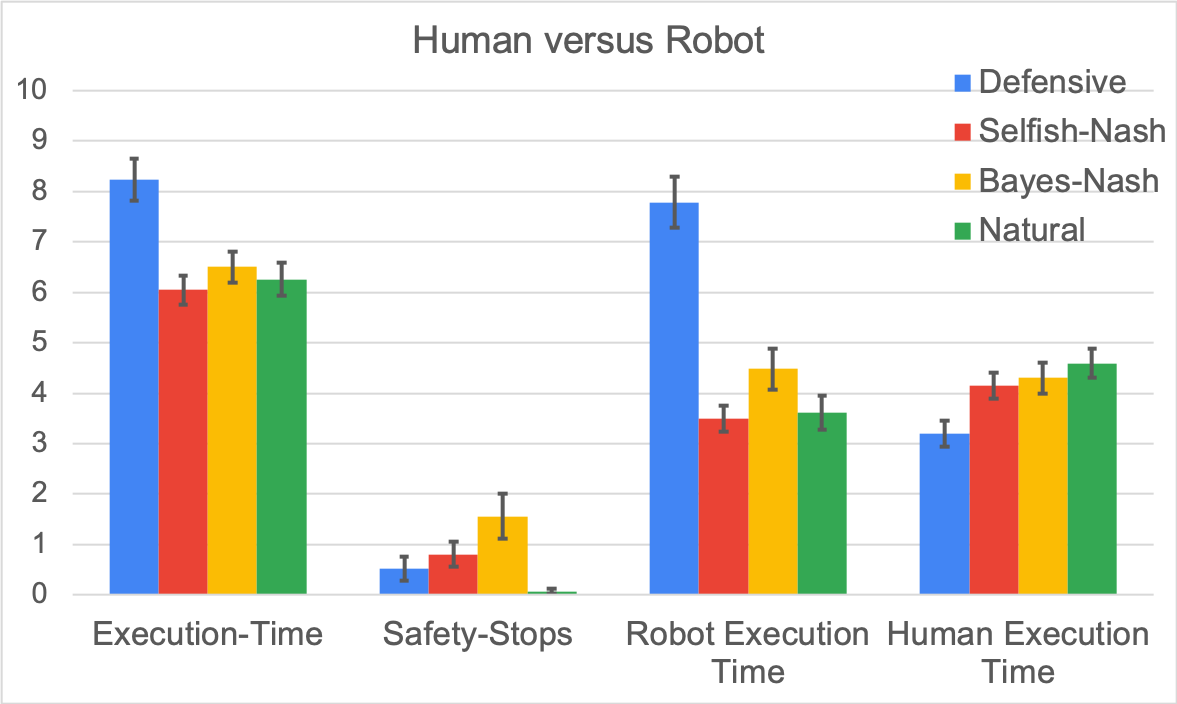}
  \caption{Human-Robot Study. Defensive and Selfish are baselines, Bayes-Nash is our approach and the Natural agent refers to the human-human study results where the participants acted naturally. Error bars represent SEM.}
    \label{fig:result_hr}
\end{figure}

\subsection{Results}
A total of 6 participants were recruited from a university campus and were randomly assigned to one possible combination of the experimental conditions. 
Figure \ref{fig:result_hr} compares the performance of the three agents. We also compare them to a naturally acting human by including the results of two naturally acting humans from the human-human study. The total task completion time was the longest for the Defensive robot while the other three agents were similar but significantly faster confirming hypothesis \textbf{H2}. The Defensive robot also had the longest robot task execution time but also led to the shortest time for the human. Bayes-Nash was the least safe and Natural the most, the selfish and defensive conditions had similarly small safety stops, contrary to hypothesis \textbf{H1}.

\subsection{Analysis} It took the Defensive agent 36.2\% more time to complete the entire task than the Selfish agent. These results confirm that the Defensive agent acts in an overly cautious manner. When comparing with the naturally-acting human we also noticed that the task completion time for the Bayes-Nash approach performs almost equally well. However, the safety stops results are surprising in two respects: (1) the higher number of safety stops for Bayes-Nash as compared to Natural and other conditions, (2) the much fewer safety stops for all conditions when compared to the simulated human study. We also noticed the latter in the human-human study, indicating that people are better at avoiding collisions as compared to the robots. We explore potential causes for these findings in the next section. 

\section{Discussion, Limitations and Future Work}
Figure \ref{fig:results_hh} shows that the Bayes-Nash approach and the naturally-acting human are similar in their ability to adapt effectively to different personalities.
However, in the human-robot study, Bayes-Nash had the most safety stops which contradicts \textbf{H1}. We believe there are two potential explanations. 

\begin{figure}[]
  \centering
  \includegraphics[width=0.9\linewidth]{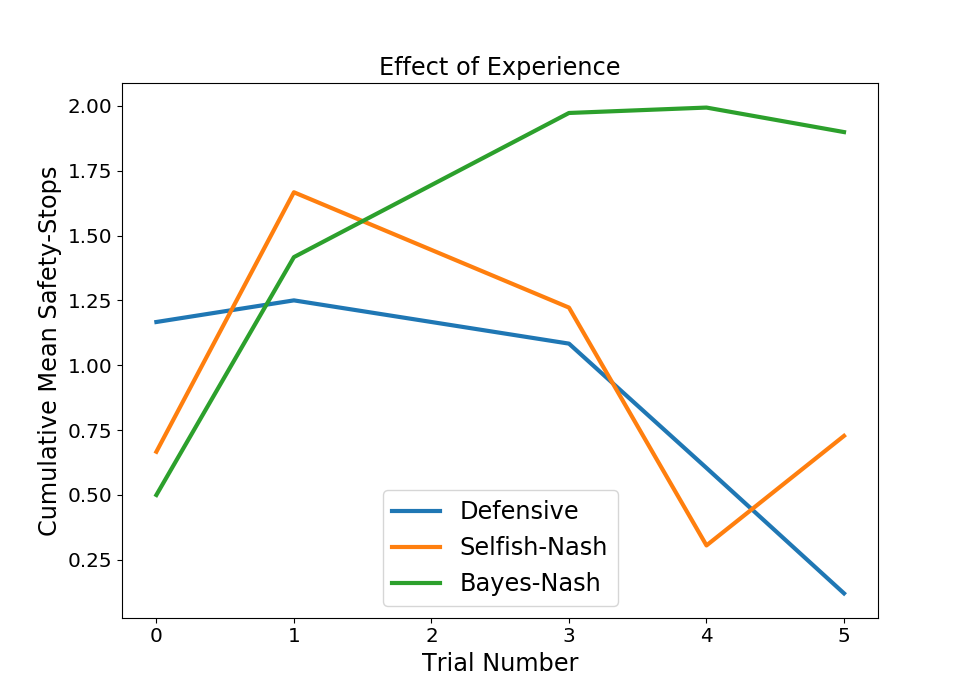}
  \caption{Effect of experience in the Human-Robot study. It shows the cumulative safety stops for the robot averaged over the trials. For the Defensive and Selfish-Nash conditions, the number of stops decreases with more trials, indicating that the human learns to avoid collision. However, for Bayes-Nash, the safety stops first increase and then remain constant, perhaps due to the difficulty in adapting to an agent whose behavior is not fixed.}
    \label{fig:learn_effect}
\end{figure}

First, due to human learning effects. Fig. \ref{fig:learn_effect} shows that the number of safety stops decrease with more trials for both the Defensive and Selfish-Nash conditions. This indicates that the user might have learned a collision-avoidant response to those agents over time. As for the Bayes-Nash condition, the safety stops first increase over trials and then remain constant. This might be because those two strategies had a somewhat fixed behavior that the human could easily adapt to. For example, if the robot moves to the goal without consideration of the human's presence every time, the human will learn that her optimal response is to wait for the robot initially. This is similar to the observation from~\cite{sadigh2016planning} where the robot directly influenced human behavior. Although this led to better performance, in our scenario, it is questionable whether this fixed behavior will be desirable from a robot collaborator in the real-world. 
Second, due to the mismatch in human-robot speed. Although the maximum arm velocities were the same for both agents, the robot was able to act faster than the human. In the Selfish condition, after a couple of trials, people might have realized that it was easier to complete the task by waiting for the robot. We need further experiments that control for these variables to confirm these and plan for it in future work. 

Although our approach can generalize to more agents, it was specifically designed for addressing scenarios involving human-robot cohabitation where usually only two agents are present~\cite{nikolaidis2015efficient,gabler2017game}. The time complexity of our algorithm is exponential in the number of agents, so can become intractable for large numbers of them. However, this may not be a limitation in real-world scenarios, since, even in cases where many agents are present (\eg, driving on a highway), we only need to consider the interactive influence of a few close-by cars to generate human-like behavior~\cite{schwarting2019social}. In the future we would like to test it in the presence of more agents. 

The primary contribution of our work is in developing a novel game-theoretic approach for HRI tasks. We also instituted a pilot study to validate this methodology and present descriptive statistics of the results. However, the small sample size did not allow us to run significance tests for the human experiments. Our plans for future work include a larger HRI study to provide evidence for generalization to a broad population.

\section*{Acknowledgments}
We would like to thank Mustafa Mukadam for discussions leading to the initial idea of this approach. We would also like to acknowledge Himanshu Sahni, Yannick Schroecker and Sangeeta Bansal for helpful discussions. 

\bibliographystyle{plainnat}
\bibliography{references}

\end{document}